%% file: main.tex
\documentclass[sigconf,nonacm]{acmart}

\pdfoutput=1
\usepackage{graphicx}
\graphicspath{ {./img/} }
\usepackage{multirow}
\usepackage{xcolor}
\usepackage{numprint}
\usepackage[misc]{ifsym}
\usepackage{tabularx}
\usepackage{subcaption}
\usepackage{float}

\AtBeginDocument{%
  \providecommand\BibTeX{{%
    \normalfont B\kern-0.5em{\scshape i\kern-0.25em b}\kern-0.8em\TeX}}}

\begin{document}

\title{{H}ow {C}ontentious {T}erms {A}bout {P}eople and {C}ultures are {U}sed in {L}inked {O}pen {D}ata}

\author{Andrei Nesterov}
\email{nesterov@cwi.nl}
\orcid{0000-0001-9697-101X}
\affiliation{%
  \institution{Centrum Wiskunde \& Informatica}
  \streetaddress{Science Park 123}
  \city{Amsterdam}
  \country{The Netherlands}
  \postcode{1098 XG}
}

\author{Laura Hollink}
\email{hollink@cwi.nl}
\orcid{0000-0002-6865-0021}
\affiliation{%
  \institution{Centrum Wiskunde \& Informatica}
  \streetaddress{Science Park 123}
  \city{Amsterdam}
  \country{The Netherlands}
  \postcode{1098 XG}
}

\author{Jacco van Ossenbruggen}
\email{jacco.van.ossenbruggen@vu.nl}
\orcid{0000-0002-7748-4715}
\affiliation{%
  \institution{VU University Amsterdam}
  \streetaddress{De Boelelaan 1105}
  \city{Amsterdam}
  \country{The Netherlands}
  \postcode{1081 HV}
}

\renewcommand{\shortauthors}{Nesterov et al.}

\begin{abstract}

Web resources in linked open data (LOD) are comprehensible to humans through literal textual values attached to them, such as labels, notes, or comments.
Word choices in literals may not always be neutral.
When outdated and culturally stereotyping terminology is used in literals, they may appear as offensive to users in interfaces and propagate stereotypes to algorithms trained on them.
We study how frequently and in which literals contentious terms about people and cultures occur in LOD and whether there are attempts to mark the usage of such terms.
For our analysis, we reuse English and Dutch terms from a knowledge graph that provides opinions of experts from the cultural heritage domain about terms' contentiousness.
We inspect occurrences of these terms in four widely used datasets: Wikidata, The Getty Art \& Architecture Thesaurus, Princeton WordNet, and Open Dutch WordNet.
Some terms are ambiguous and contentious only in particular senses.
Applying word sense disambiguation, we generate a set of literals relevant to our analysis.
We found that outdated, derogatory, stereotyping terms frequently appear in descriptive and labelling literals, such as preferred labels that are usually displayed in interfaces and used for indexing.
In some cases, LOD contributors mark contentious terms with words and phrases in literals (implicit markers) or properties linked to resources (explicit markers).
However, such marking is rare and non-consistent in all datasets.
Our quantitative and qualitative insights could be helpful in developing more systematic approaches to address the propagation of stereotypes via LOD.

\end{abstract}


\settopmatter{printfolios=true} 
\maketitle

\noindent \textbf{Disclaimer.}
In this paper, contentious words and phrases presented in \textit{``quotation marks and italicised''} can be derogatory and offensive.
They are provided solely as illustration of the research and do not reflect the opinions of the authors or their organisations.

\section{Introduction}\label{sec:intro}
\input{src/intro.tex}

\section{Related Work}\label{sec:related}
\input{src/related_work.tex}

\section{Data}\label{sec:data}
\input{src/data.tex}

\section{Approach}\label{sec:approach}
\input{src/approach.tex}

\section{Results}\label{sec:results}
\input{src/results.tex}

\section{Conclusion}\label{sec:conclusion}
\input{src/conclusion.tex}


\bibliographystyle{ACM-Reference-Format}
\bibliography{bibliography}

\appendix
\input{src/appendix.tex}

\end{document}

%% file: src/intro.tex
Developers of knowledge graphs and linked open data (LOD) describe machine-readable web resources with literal values to make them also human-readable.
Labels, notes, comments, and other literals contain information in natural language, which can be presented to users in interfaces.
Some data providers state in their guidelines that developers and contributors should be aware of their word choices when describing resources to stay impartial.
For example, Wikidata recommends to its contributors to ``keep descriptions neutral by avoiding opinionated or biased terms''.\footnote{\url{https://www.wikidata.org/w/index.php?title=Help:Description\&oldid=1889469468}}
The guidelines of The Getty Art \& Architecture Thesaurus (AAT), a controlled vocabulary for the cultural heritage domain, instruct its editors to ``be objective'' and avoid language that expresses cultural biases or ``may be considered offensive by groups of people''.\footnote{\url{https://www.getty.edu/research/tools/vocabularies/guidelines/aat_3_4_scope_note.html}. See paragraph 3.4.1.5.11. Last accessed on 13.11.2023}
Our study demonstrates, however, that outdated and derogatory words, colonial categories, racial slurs, and other contentious terms (for example \textit{``coolie''}, \textit{``colored''}, \textit{``hottentot''}, \textit{``mongoloid''}, \textit{``mulatto''}, \textit{``negro''}, \textit{``transvestite''}) are still being used to describe resources about (historically) marginalised people and cultures in widespread linked open datasets.
If such stereotyping language is left unexamined, there is a risk of further propagation of stereotypes in user interfaces \cite{BiasedLanguage} and ML algorithms that use literals as training data \cite{SurveySocialBiasKG}.

We collect empirical evidence on how contentious terms about people and cultures occur in LOD to illustrate the extent of their usage in literals and provide insights into the existing practices of handling potentially stereotyping language.
For our study, we adopted a list of English and Dutch terms from a knowledge graph of contentious terminology \cite{ContentiousTerminologyKG}, which describes terms related to historically marginalised peoples (\textit{``gypsy''}, \textit{``eskimo''}, \textit{``indigenous''}), colonial territories (\textit{``Batavia''}), traditions (\textit{``headhunter''}), and other categories.
This knowledge graph is based on the publication ``Words Matter'' \cite{WordsMatter} produced by professionals from the cultural heritage domain, who recommend whether certain terms referring to groups of people and cultures should be avoided, marked and explained, or replaced with appropriate synonyms in cultural heritage datasets.
This expert knowledge allows us to circumvent judging ourselves whether terms are contentious or not.
We operationalize the definition ``contentious terms'' to refer to the terms we extract from the knowledge graph, in which these terms were explicitly defined as contentious.

Our analysis focusses on four datasets with English and Dutch literals: Wikidata, AAT, and the lexical databases Princeton WordNet and Open Dutch WordNet.
These datasets encompass a variety of domains and applications, have different structure and level of curation, which can enrich our insights.

We answer two research questions. \textbf{RQ1}: In which literals and how often are contentious terms used in LOD datasets? \textbf{RQ2}: Do contentious terms in literals have any markers of their contentiousness and if so, what are these markers and how are they given: implicitly (in text of literals next to contentious terms) or explicitly (via specific properties)?

From the four LOD datasets, we extract occurrences of contentious terms in the literals used for labelling (such as ``preferred label'') as well as descriptive literals (for example, ``notes'').
Some contentious terms are ambiguous.
This leads to retrieving literals in which the found terms are used in a non-contentious sense.
For example, the ambiguous term \textit{``primitive''} is used in such literals as ``primitive data type'' and \textit{``primitive society''}.
According to the reused knowledge graph, only the latter literal mentions the term in the contentious sense.
We apply a word sense disambiguation approach that enables us to gather a subset of resources with more relevant literals mentioning terms in their contentious senses.
Additionally, we analyse a smaller, highly reliable set of resources, which were collected manually to ensure that contentious terms in their literals are the closets in meaning and scope to the terms from the knowledge graph.

To identify implicit and explicit markers of contentiousness, we first inspect literals and properties of the manually selected resources.
Then, we find similar markers automatically in all the resources we extracted from the four datasets as well as the resources with disambiguated literals.

The contributions of this paper are the following:
\begin{enumerate}
    \item Our quantitative findings prove that most of the investigated English and Dutch contentious terms are still being used frequently in LOD datasets.
    They occur in descriptive property values (for example notes, descriptions, definitions) as well as in labels (preferred or alternative).
    There are implicit and explicit markers of contentiousness found in all datasets, however, they are rarely used.
    \item Our qualitative results reveal problematic cases when literals, including preferred labels, communicate stereotypes about people and cultures with outdated or derogatory terms and slurs.
    We describe different ways of how contentious terms are currently being marked in LOD, however, the usage of these markers is not consistent.
\end{enumerate}  

Our findings, including the datasets of resources with disambiguated literals and contentiousness markers, are meant as a starting point for researchers, data curators, and developers to design more systematic and reusable approaches of addressing potentially stereotyping terminology on the Web.

%% file: src/related_work.tex
\paragraph{Cultural Bias in Knowledge Graphs}

Different types of bias have been investigated in knowledge graphs \cite{DebiasingKnowledgeGraphs,MLmeetsSW}. 
Bias may spread to applications powered by knowledge graphs, such as search engines or question-answering systems \cite{QuantifyingTheGap,ImplicitBiasCrowdsourcedKGs}, subsequently influencing end-users' perception of represented information \cite{DebiasingKnowledgeGraphs,RaceAndCitizenshipWikidata}.
Knowledge graph embeddings capture bias \cite{MeasuringSocialBiasinKGEmbeddings} and propagate it to algorithms, for example, recommender systems \cite{BiasKGMovies}.
Existing empirical research on bias in knowledge graphs focusses mainly on data imbalances in representation of gender \cite{QuantifyingTheGap,MonitoringGenderGapWikidata}, citizenship and race in Wikidata \cite{RaceAndCitizenshipWikidata}.
In our research, we analyse literals that may contain bias expressed with particular contentious terms.
This type of bias differs from the data imbalances bias.
It is ``bias in a cultural context'' \cite{DebiasingKnowledgeGraphs} that propagates prejudices and stereotypes against people and cultures.
This bias may come from, but is not limited to, ``socio-cultural'' or ``political and religious'' factors \cite{ExplorationCognitiveBiasOntologies}.

There are different stages on which cultural bias can infiltrate knowledge systems, for example during the creation of ontologies or data \cite{SurveySocialBiasKG}.
We focus on the stage of literals creation and gather empirical evidence on whether, and if so, how LOD developers address cultural contentiousness.

\paragraph{The Role of Literals in Linked Open Data}

Literals with natural language have numerous purposes in LOD.
They make data more accessible to users providing human-readable information, allowing indexing \cite{Lotus} and search in natural language \cite{FREyA}.
Besides the user-experience purposes, literals serve as background information in approaches of entity linking \cite{SurveyEnglishEL} and ontology alignment \cite{StringSimilarityMetrics}, including approaches with knowledge graph embeddings \cite{ComprehensiveSurveyKGEmbeddings}.

Prior research into natural language literals primarily relates to the quality of LOD.
A taxonomy to analyse the quality of literals in linked data is presented in \cite{LiterallyBetter}, in which the quality of ``textual strings'' is viewed mainly as compliance to the standards of language tagging.
The influence of labelling practices on the accessibility of LOD to applications and users is studied in \cite{TheHumanFace} and \cite{LabelsWebData} across several metrics: such as completeness (if all entities carry human-readable information), efficiency of querying triples with literals, unambiguity of labelling properties, and multilinguality of language tags.
For the purpose of calculating such metrics, those studies provide the distribution of properties used for labelling in different datasets (including Wikidata in \cite{TheHumanFace}).

Terms used in a knowledge graph inform about its scope, as it is demonstrated in the study, in which terms (labels) of resources are extracted and visualised to profile cultural heritage knowledge graphs \cite{TTProfiler}.
In our work, besides labels, we also extract longer descriptive literals to further investigate how contentious terms occur in knowledge graphs.

%% file: src/data.tex
\begin{table}[t]
\caption{N literals in the analysed LOD-datasets by properties}
\label{tab:lod-resources}
\centering
\resizebox{\columnwidth}{!}{%
\begin{tabular}{|l|l|ll|}
\hline
\multirow{2}{*}{\textbf{Dataset}} & \multirow{2}{*}{\textbf{Property}} & \multicolumn{2}{l|}{\textbf{\# literals}} \\ \cline{3-4} 
                                    &                                               & \multicolumn{1}{l|}{\textbf{EN}}          & \textbf{NL}          \\ \hline
\multirow{3}{*}{Wikidata}           & skos:prefLabel                                & \multicolumn{1}{l|}{86,397,295}             & 62,964,028             \\ \cline{2-4} 
                                    & skos:altLabel                                 & \multicolumn{1}{l|}{7,324,345}              & 1,357,679              \\ \cline{2-4} 
                                    & schema:description                            & \multicolumn{1}{l|}{83,565,644}             & 78,592,681             \\ \hline
\multirow{5}{*}{AAT}          & xl:prefLabel\textbackslash xl:literalForm    & \multicolumn{1}{l|}{49,892}                & 43,788                \\ \cline{2-4} 
                                    & xl:prefLabel\textbackslash rdfs:comment      & \multicolumn{1}{l|}{130}                  & 42                   \\ \cline{2-4} 
                                    & xl:altLabel\textbackslash xl:literalForm     & \multicolumn{1}{l|}{114,119}               & 28,992                \\ \cline{2-4} 
                                    & xl:altLabel\textbackslash rdfs:comment       & \multicolumn{1}{l|}{342}                  & 5                    \\ \cline{2-4} 
                                    & skos:scopeNote\textbackslash rdf:value       & \multicolumn{1}{l|}{42,851}                & 34,280                \\ \hline
\multirow{3}{*}{PWN}  & ontolex:writtenRep                            & \multicolumn{1}{l|}{207,272}               & \multirow{3}{*}{N/A} \\ \cline{2-3}
                                    & wn:definition (``Definition'')                 & \multicolumn{1}{l|}{117,791}               &                      \\ \cline{2-3}
                                    & wn:definition (``Examples'')                   & \multicolumn{1}{l|}{32,990}                &                      \\ \hline
\multirow{4}{*}{ODWN} & Lemma writtenForm                             & \multicolumn{1}{l|}{\multirow{4}{*}{N/A}} & 90,897                \\ \cline{2-2} \cline{4-4} 
                                    & Sense definition                              & \multicolumn{1}{l|}{}                     & 50,021                \\ \cline{2-2} \cline{4-4} 
                                    & SenseExamples & \multicolumn{1}{l|}{}                     & 31,600                \\ \cline{2-2} \cline{4-4} 
                                    & Synset Definition gloss                       & \multicolumn{1}{l|}{}                     & 32,098                \\ \hline
\end{tabular}
}
\end{table}

As a starting point of our research, we reuse a knowledge graph of contentious terminology \cite{ContentiousTerminologyKG} based on expert knowledge from the Words Matter publication \cite{WordsMatter}, which we call the Words Matter KG.
There are 75 English and 82 Dutch contentious terms described in the Words Matter KG, each term has a link to explanations and suggestions on its usage.
In some cases, suggestions for contentious terms provide synonymous terms which are judged to be more appropriate.
Apart from the terms, we also reuse links between them and closely related resources from external LOD-datasets provided in the Words Matter KG.
Contentious terms are represented as SKOS-XL labels and connected to 58 English and 63 Dutch entities in Wikidata, 37 English and 27 Dutch concepts in AAT, and 81 synsets in PWN via the property \textit{skos:relatedMatch}.
For example, the Wikidata entity Q3254959\footnote{\url{https://www.wikidata.org/w/index.php?title=Q3254959&oldid=1923044725}}
is about \textit{``race''} in the sense of human categorization by physical features, which is linked to the term \textit{``race''} in the Words Matter KG, because it is related to the term's meaning.
Such related resources were collected by human annotators.

We extend the list of terms with their inflected forms.
For example, the canonical term \textit{``aboriginal''} also has the plural form \textit{``aboriginals''}.
With canonical and inflected forms, the resulting list includes 154 English and 242 Dutch terms.
We retrieve literals containing terms from this list in the following LOD-datasets:

\begin{description}
    \item[Wikidata], one of the largest user-generated knowledge bases on the Web with 12,5 billion triples,\footnote{\url{https://lod-cloud.net/dataset/wikidata}. Last accessed on 30.08.2023} in English and Dutch;
    \item[The Getty Art \& Architecture Thesaurus (AAT)], a controlled vocabulary\footnote{\url{https://www.getty.edu/research/tools/vocabularies/aat/about.html}.\\Last accessed on 13.11.2023} that serves as a reference for cultural heritage institutions,\footnote{\url{https://pro.europeana.eu/post/europeana-enriches-its-data-with-the-art-and-architecture-thesau}. Last accessed on 13.11.2023} in English and Dutch;
    \item [Princeton WordNet (PWN)] \cite{PWN}, a lexical database for English, which is frequently used in ML tasks, such as word sense disambiguation \cite{SenseAnnotatedCorpora};
    \item [Open Dutch WordNet (ODWN)] \cite{ODWN}, a lexical database for Dutch, which is a part of the larger Open Multilingual WordNet\footnote{\url{https://omwn.org/}. Last accessed on 13.11.2023} and other dictionaries, for example BabelNet.\footnote{\url{https://babelnet.org/}. Last accessed on 13.11.2023}
\end{description}

We extract literals from property values that give main names for resources (labels) as well as descriptive properties.
Table \ref{tab:lod-resources} presents an overview of the selected properties in each dataset and overall counts of their literal values.\footnote{Wikidata counts: \url{https://www.wikidata.org/w/index.php?title=User:Mr.\_Ibrahem/Language\_statistics\_for\_items\&direction=next\&oldid=1816384208}}
Wikidata uses redundant properties for defining labels: we use \textit{skos:prefLabel} ignoring the \textit{rdfs:label} and \textit{schema:name} properties with identical values.
Alternative labels and descriptions are defined using \textit{skos:altLabel} and \textit{schema:description}.
Similar to Wikidata, AAT provides preferred and alternative labels for concepts but using SKOS-XL,\footnote{\url{https://www.w3.org/TR/skos-reference/skos-xl.html}. Last accessed on 13.11.2023} which represents labels as URIs with literal values given via \textit{xl:literalForm}.
In a few cases, there is additional textual information about labels contained in the values of \textit{rdfs:comment}.
We also search for terms in literals of the \textit{skos:scopeNote} property (via the \textit{rdf:value} path).

In PWN (version 3.1), we search contentious terms in lemmas, definitions, and examples of synsets.
Lemmas represent canonical forms of words similar to dictionary entries, and synonymous lemmas are grouped in synsets. 
The RDF representation models lemmas with the property \textit{ontolex:writtenRep} and synset definitions with the property \textit{wn:definition}.
Examples are not represented separately, but included in definition texts and separated with double quotes.
In our analysis, we separate definitions from examples. 

ODWN (version 1.3) is available only in XML format.
Although, it is not released as LOD, we have included ODWN in the study, because it is the Dutch language counterpart of PWN and it constitutes other WordNets and dictionaries.
We provide the names of selected XML tags with literals in Table~\ref{tab:lod-resources}.
Lemmas in ODWN are contained in the attribute value of the \textit{Lemma writtenForm} tag, which is a child of \textit{LexicalEntry}.
Lexical entries have child elements \textit{Senses}, which are connected to synsets. 
Unlike in PWN, there are two types of definitions in ODWN: definitions attached to senses (\textit{Sense definition}) and definitions of synsets (the attribute value of \textit{Definition gloss} with the parent \textit{Definitions}).
Examples are given for senses and explicitly separated from definitions with the tag \textit{SenseExamples}; we used its both child tags \textit{textualForm} and \textit{canonicalForm}.
For ODWN, we ignored the English literals.

Note that while English and Dutch are the two languages with the largest number of preferred labels and descriptions in Wikidata; and the largest number of preferred labels, alternative labels, and scope notes in AAT; there are significantly fewer Dutch alternative labels than English in both datasets (see Table \ref{tab:lod-resources}).

%% file: src/approach.tex
\begin{figure}[ht]
  \centering
  \includegraphics[width=\linewidth]{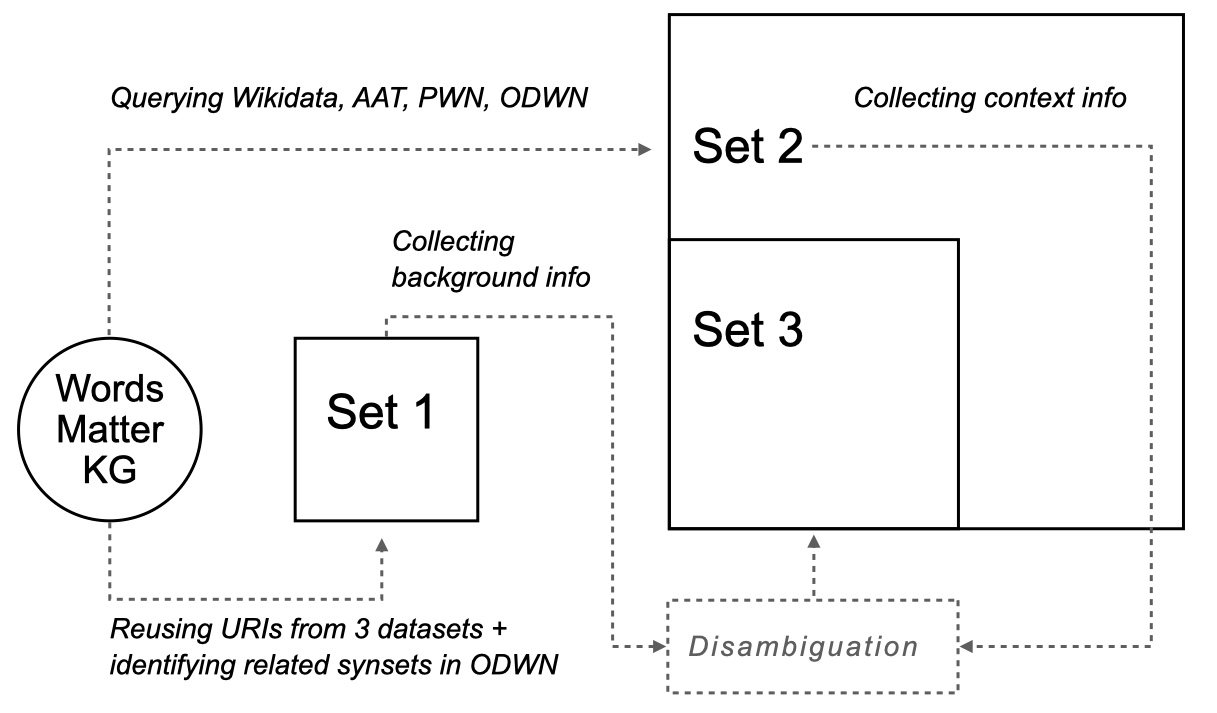}
  \caption{Set 3 with disambiguated literals is a subset of the largest Set 2 with all retrieved literals.
  Background information taken from the literals in Set 1 is used in disambiguation.}
  \label{fig:approach_diagram}
  \Description{}
\end{figure}

We construct three sets of literals.
In this section, we first explain how we reuse related resources from the Words Matter KG to create Set 1. 
Then, we describe the process of querying English and Dutch contentious terms in 4 datasets (Set 2).
The literals from the query results required disambiguation.
We explain and evaluate our disambiguation approach, which helped us to generate a subset of literals more relevant to our analysis, which we call Set 3.
Figure \ref{fig:approach_diagram} illustrates the relationships between the Words Matter KG and the three sets.
Lastly, we explain how implicit and explicit markers of contentiousness were identified in each set.

\subsection{Reusing Related Resources (Set 1)}\label{approach_set1}

From the Words Matter KG, we take URIs of the related resources in Wikidata, AAT, and PWN.
Since links between contentious terms and ODWN were not yet available, we add 65 links between contentious terms and lexical entries in ODWN using the same guidelines as in \cite{ContentiousTerminologyKG}. 
By retrieving the literals of all related resources, we construct Set 1.
The results from this set are highly reliable, because the manual identification of resources aimed at ensuring that contentious terms in their literals are closest in meaning and scope to the terms from the Words Matter KG.

Besides analysing Set 1, we also use it as background information of contentious terms during disambiguation to construct Set 3 (see Step 1 in Subsection \ref{approach_set3}).

\subsection{Querying Literals in LOD (Set 2)} \label{approach_set2}

From each LOD dataset, we retrieve resources that have contentious terms in the literal values of the labelling and descriptive properties.
In this subsection, we explain how we queried the literals in each dataset.
The retrieval pipelines are packaged in an open license Python module which we call \textit{LODlit}.\footnote{\url{https://github.com/mnn-001-p/LODlit}}
It serves as a tool to reproduce our results and can be reused for other tasks.

\paragraph{Wikidata}

We use the MediaWiki Action API\footnote{\url{https://www.mediawiki.org/wiki/API:Main_page}. Last accessed on 13.11.2023} to retrieve Wikidata entities.
The API limits requests to {10K} results.
Because some terms yield more search results, we use the top {10K} ranked by incoming links.
The search results were dominated by instances of a few over-represented categories likely irrelevant to our analysis.
Therefore, we take three additional steps to filter these out and increase the relevance of the results.
First, we exclude entities belonging to 10 selected categories.
Excluded are, for example, instances and subclasses of ``scholarly article'' (Q13442814) and ``taxon'' (Q16521).
Second, we exclude entities with the phrases ``scholarly'' or ``scientific article'' in their descriptions, because many scientific articles are not categorised as such with properties.
Third, we filter out entities for which the query term appears as a person's name.
For example, the terms ``black'' and ``page'' appear in the names Jack Black and Brian S. Page. 
For this purpose, we filter out entities if the value of P31 (``instance of'') is Q5 (``human'') AND if the preferred label contains a capitalised query term.
For the complete list of excluded categories and the breakdown of search results before and after filtering, we refer to the research documentation.\footnotemark[11]
Wikidata resources were retrieved on 31.01.2023.

\paragraph{AAT}

The Getty Research Institute provides a SPARQL endpoint, which we use to construct English and Dutch subgraphs of the AAT.\footnote{\url{http://vocab.getty.edu/sparql}, both subgraphs were retrieved on 12.09.2022}
Then, we query the subgraphs for concepts that have a contentious term in their literal values.

\paragraph{PWN}

From Princeton WordNet 3.1, we retrieve all synsets with contentious terms using Python's NLTK package.

\paragraph{ODWN}

We query Open Dutch WordNet using the dedicated Python module.\footnote{\url{https://github.com/cltl/OpenDutchWordnet}. Last accessed on 13.11.2023}
Because not all properties could be queried with the native module functions, some changes were made, which we documented and archived.\footnote{\url{https://github.com/mnn-001-p/LODlit}}

\paragraph{Aggregating Search Results}

We count the number of occurrences (hits) of each contentious term per dataset, language, and property.
For example, the query ``slave'' has 21 hits in English literals of AAT appearing 2 times in preferred labels, 17 times in alternative labels, and 2 times in scope notes.
The plural form ``slaves'' has 19 hits in English AAT.
For the analysis, we group terms by their canonical form, so the term ``slave'' has 40 hits in total. 
A resource may have multiple values of one property that mention the same contentious term.
For example, a resource has two alternative labels \textit{``slave owner''} and \textit{``slave master''}.
We count these as separate hits.
If a single literal mentions the same contentious term multiple times, we count this as one hit (for example, the difinition from PWN: \textit{``relating to or involving slaves or appropriate for slaves or servants''}).

\subsection{Disambiguating Literals (Set 3)}\label{approach_set3}

Some contentious terms appear in LOD datasets in more than one sense.
This may lead to a large number of irrelevant results.
For example, the term ``colored'' appears in more than 10 thousand literals we retrieved.
Many of these are not in the contentious sense ``non-white people'' but rather in the sense of ``having colour, colourful''.
To construct a set with more relevant literals mentioning contentious terms, we perform word sense disambiguation (WSD).

Before designing our disambiguation process, we tested off-the-shelve entity linking tools, such as Babelfy \cite{Babelfy} and Falcon \cite{Falcon}, which link concepts mentioned in an input text to concepts from knowledge bases.
However, in our case, these tools did not resolve the challenges associated with our datasets' size (limits of the Babelfy API), the language and varying length of literals (Falcon is designed for short English texts).

To disambiguate English and Dutch literals en masse, we set up a WSD process based on the approach of calculating similarity between the word vectors of a target term's context and its sense inventory \cite{WSD_CS_word2vec}.
Instead of a sense inventory, we use terms' background information specifying their contentious senses.

Our WSD process includes four steps.
On Step 1, we experimented with an alternative source of background information – descriptions of contentious terms provided in the Words Matter KG.
We also experimented with common token overlap between background and context information of a term instead of calculating cosine similarity scores between word vectors.
The experiments showed that the steps we describe below produced better results.\footnote{\url{https://github.com/mnn-001-p/LODlit}}

\paragraph{Step 1. Collecting Background Information.}
For each contentious term, we collect background information from the literals of its related resources in Set 1.
For Wikidata resources, if a contentious term occurs in an entity's labels, we extend the background information with the literals of the properties ``instance of'' (P31) and ``subclass of'' (P279).
For ODWN resources, if a contentious term was found in the \textit{Synset Definition}, we use only that definition and the \textit{Lemma writtenForm} of the synset as background information, but not \textit{Sense definitions} or \textit{Sense examples} of that lemma, because they are specific to lemmas.
The background information text is then converted to a bag-of-words (BoWs) by tokenising, lemmatising, and removing non-word characters, digits, and stop words in English and Dutch.
Tokens of fewer than 3 characters are discarded.

\paragraph{Step 2. Collecting Context Information.}
For each retrieved resource in Set 2, we collect context information from the literals of labelling and descriptive properties.
Context information is converted into a BoW in the same way as background information.

\paragraph{Step 3. Using Pre-trained Word Vectors.}
For every token in the BoWs of background and context information, we infer word vectors from the pre-trained open-license language models by Spacy.\footnote{\url{https://spacy.io/models}. Last accessed on 13.11.2023}
We use the \textit{en\_core\_web\_lg} and \textit{nl\_core\_news\_lg} models for English and Dutch tokens respectively.
Subsequently, each BoW is represented as an average of all token vectors it contains.

\paragraph{Step 4. Ranking by Cosine Similarity Scores.}
We calculate cosine similarity (CS) between the context vector and the corresponding background vector for each resource in Set 2. 
The higher the CS, the more likely the resource mentions the contentious term in the sense intended in the Words Matter KG.
We filter out resources that are likely not relevant setting a CS threshold to 0.5.
To construct Set 3 with disambiguated literals, we rank all resources by CS and take maximum 10 resources with the highest scores (top-10) per term's canonical form.
Some groups of terms had a small number of resources, which led to fewer than 10 entities per canonical form added to the set.

\paragraph{Evaluation}

We annotated a sample of the resulting set.
Two annotators, who are co-authors of this paper, checked whether the sense of the contentious term mentioned in the resource is similar to the sense of the same term in the provided background information.
We used a stratified sample to ensure that both common and rare terms were included in the evaluation.
We divided (canonical) terms into quartiles based on the number of search results they yielded. 
For example, rare terms, such as \textit{``coolie''} or \textit{``half-blood''}, are in the first quartile, while terms with a lot of results, such as \textit{``black''}, are in the fourth quartile.
Then, we drew 10 random resources from each quartile, which resulted in 40 resources per dataset per language.
Thus, each annotator checked 240 resources for the four datasets (two of which are in two languages).

Inter-rater agreement was high with a Krippendorff’s $\alpha$ of 0.80.
Annotators deemed 72\% of the resources in the sample relevant.
In addition, as a form of common sense evaluation, we checked whether Set 3 included the closely related resources from Set 1.
Intuitively, these resources should be included, because they would have a high CS score.
For Dutch, all related resources were included. 
For English, all resources from PWN were included; from AAT, all but 1 were included; and for Wikidata, all but 8 were included.

\subsection{Identifying Markers of Contentiousness}\label{approach_markers}

Some resources contain information about contentiousness of terms used in their literals.
We call such information ``markers of contentiousness'', which can be implicit and explicit.
Implicit markers are words and phrases found in literals alongside contentious terms in the same or different property values of a resource.
As an example, the synset \textit{``fagot.n.01''} from PWN with the term \textit{``queer''} in lemmas is defined as ``offensive term for a homosexual man''.
The word ``offensive'' is an implicit marker.
Properties and relations between resources indicating that a resource uses contentious terms are explicit markers.
For example, the property P31 (``instance of'') in Wikidata links the entity Q1135775 (\textit{``redneck''}) to another entity Q545779 (``pejorative'').
The principal difference between implicit and explicit markers is that the latter have machine readable URIs or tags opposed to implicit markers requiring textual search.

We identify implicit and explicit markers of contentiousness in all sets.
First, we manually collect both the words associated with implicit markers (for example, ``offensive'', ``derogatory'', ``outdated''), properties of explicit markers present in Set 1, and additional markers we found while querying datasets.
Second, we automatically collect markers in Sets 2 and 3: for implicit markers we use string matching with regular expressions, for explicit – SPARQL (in Wikidata and AAT) and custom Python functions (in PWN and ODWN).

Besides the markers of contentiousness, we search for words and phrases that were suggested to be used instead of contentious terms in the Words Matter KG.
We find suggestions in Sets 1 and 3 using fuzzy string matching.
The usage of these suggestions together with contentious terms in the same or different property values can be also indicative of terms’ contentiousness.

%% file: src/results.tex
For each of the 3 sets, we report how frequently contentious terms occur in all datasets' property values (Table \ref{tab:hits_all_levels}).
We answer why some datasets, properties, and terms stand out with a large number of hits, check whether synonyms of contentious terms are also used in literals, and provide examples of potentially problematic situations.
Separately, we report on which implicit and explicit markers of contentiousness we identified, how frequently they are used, and what they indicate (Table \ref{tab:markers_types}).
For Set 2 with all query results, we present figures with the most frequent contentious terms in Appendix \ref{appx_set_1_figures}.
Additionally, there are interactive figures with distribution of hits by properties for each contentious term in the online appendix to this paper.\footnote{\url{https://github.com/mnn-001-p/LODlit}}

\begin{table}
\caption{N hits of contentious terms in 3 result sets}
\label{tab:hits_all_levels}
\centering
\resizebox{\columnwidth}{!}{%
\begin{tabular}{|l|l|ll|ll|ll|}
\hline
\multirow{2}{*}{\textbf{Dataset}} &
  \multirow{2}{*}{\textbf{Properties}} &
  \multicolumn{2}{l|}{\textbf{Set 2}} &
  \multicolumn{2}{l|}{\textbf{Set 3}} &
  \multicolumn{2}{l|}{\textbf{Set 1}} \\ \cline{3-8} 
 &
   &
  \multicolumn{1}{l|}{\textbf{EN}} &
  \textbf{NL} &
  \multicolumn{1}{l|}{\textbf{EN}} &
  \textbf{NL} &
  \multicolumn{1}{l|}{\textbf{EN}} &
  \textbf{NL} \\ \hline
\multirow{3}{*}{Wikidata} &
  skos:prefLabel &
  \multicolumn{1}{l|}{84,261} &
  7,327 &
  \multicolumn{1}{l|}{364} &
  321 &
  \multicolumn{1}{l|}{44} &
  51 \\ \cline{2-8} 
 &
  skos:altLabel &
  \multicolumn{1}{l|}{18,126} &
  1,518 &
  \multicolumn{1}{l|}{306} &
  132 &
  \multicolumn{1}{l|}{67} &
  53 \\ \cline{2-8} 
 &
  schema:description &
  \multicolumn{1}{l|}{87,058} &
  33,885 &
  \multicolumn{1}{l|}{210} &
  137 &
  \multicolumn{1}{l|}{4} &
  0 \\ \hline
\multirow{5}{*}{AAT} &
  xl:prefLabel\textbackslash xl:literalForm &
  \multicolumn{1}{l|}{526} &
  295 &
  \multicolumn{1}{l|}{58} &
  55 &
  \multicolumn{1}{l|}{23} &
  25 \\ \cline{2-8} 
 &
  xl:prefLabel\textbackslash rdfs:comment &
  \multicolumn{1}{l|}{6} &
  0 &
  \multicolumn{1}{l|}{1} &
  0 &
  \multicolumn{1}{l|}{0} &
  0 \\ \cline{2-8} 
 &
  xl:altLabel\textbackslash xl:literalForm &
  \multicolumn{1}{l|}{1,428} &
  94 &
  \multicolumn{1}{l|}{155} &
  14 &
  \multicolumn{1}{l|}{60} &
  11 \\ \cline{2-8} 
 &
  xl:altLabel\textbackslash rdfs:comment &
  \multicolumn{1}{l|}{6} &
  0 &
  \multicolumn{1}{l|}{1} &
  0 &
  \multicolumn{1}{l|}{0} &
  0 \\ \cline{2-8} 
 &
  skos:scopeNote\textbackslash rdf:value &
  \multicolumn{1}{l|}{5,742} &
  3,568 &
  \multicolumn{1}{l|}{342} &
  279 &
  \multicolumn{1}{l|}{10} &
  6 \\ \hline
\multirow{3}{*}{\begin{tabular}[c]{@{}l@{}}PWN\end{tabular}} &
  ontolex:writtenRep &
  \multicolumn{1}{l|}{230} &
  \multirow{3}{*}{N/A} &
  \multicolumn{1}{l|}{124} &
  \multirow{3}{*}{N/A} &
  \multicolumn{1}{l|}{90} &
  \multirow{3}{*}{N/A} \\ \cline{2-3} \cline{5-5} \cline{7-7}
 &
  wn:definition (``Definition'') &
  \multicolumn{1}{l|}{5,148} &
   &
  \multicolumn{1}{l|}{244} &
   &
  \multicolumn{1}{l|}{11} &
   \\ \cline{2-3} \cline{5-5} \cline{7-7}
 &
  wn:definition (``Examples'') &
  \multicolumn{1}{l|}{578} &
   &
  \multicolumn{1}{l|}{126} &
   &
  \multicolumn{1}{l|}{36} &
   \\ \hline
\multirow{4}{*}{\begin{tabular}[c]{@{}l@{}}ODWN\end{tabular}} &
  Lemma writtenForm &
  \multicolumn{1}{l|}{\multirow{4}{*}{N/A}} &
  147 &
  \multicolumn{1}{l|}{\multirow{4}{*}{N/A}} &
  86 &
  \multicolumn{1}{l|}{\multirow{4}{*}{N/A}} &
  66 \\ \cline{2-2} \cline{4-4} \cline{6-6} \cline{8-8} 
 &
  Sense definition &
  \multicolumn{1}{l|}{} &
  219 &
  \multicolumn{1}{l|}{} &
  70 &
  \multicolumn{1}{l|}{} &
  2 \\ \cline{2-2} \cline{4-4} \cline{6-6} \cline{8-8} 
 &
  SenseExamples &
  \multicolumn{1}{l|}{} &
  433 &
  \multicolumn{1}{l|}{} &
  171 &
  \multicolumn{1}{l|}{} &
  38 \\ \cline{2-2} \cline{4-4} \cline{6-6} \cline{8-8} 
 &
  Synset Definition gloss &
  \multicolumn{1}{l|}{} &
  190 &
  \multicolumn{1}{l|}{} &
  62 &
  \multicolumn{1}{l|}{} &
  0 \\ \hline
\end{tabular}%
}
\end{table}

\subsection{Set 1: Closely Related Resources}\label{set1}

This set includes only resources closest in meaning to contentious terms.
There are 176 and 155 resources with English and Dutch literals, respectively, in which we found 345 English and 252 Dutch hits of contentious terms (``Set 1'' in Table \ref{tab:hits_all_levels}).
6 English and 5 Dutch terms did not have related resources in any of the datasets.
AAT had the least related resources leaving 34 English and 55 Dutch terms without links to corresponding concepts, which was unexpected, since the terms originate from the cultural heritage domain.
The terms without related resources are listed in Appendix \ref{terms_no_related_resources}.

Primarily, contentious terms in both languages are found in preferred and alternative labels of Wikidata and AAT as well as in lemmas of PWN and ODWN.
This is expected since closely related resources mention contentious terms in labels.

Resources with contentious terms in preferred labels often denote (groups of) people, for example \textit{``coolie''} (Q548135), \textit{``mulatto''} (Q191923), \textit{``transvestite''} (Q112918934), \textit{``pygmy''} (Q171927) in Wikidata, \textit{``homosexuals (people)''} (300435115), \textit{``Indiaans''} (\textit{``Indians''} in Dutch, referring to Native American people) (300017437) in AAT, \textit{``mongool''} (\textit{``mongoloid''} in Dutch, referring to people with Down syndrome) (synset eng-30-10197525-n) in ODWN.
Such preferred labels are displayed online as standard names of resources, which creates risks of propagation of stereotypes.
In some cases, the resources are described as \textit{terms}, for example Wikidata's \textit{``Colored''} (Q5149038) with the description ``Term used in the United States to describe black people''\footnote{\url{https://www.wikidata.org/w/index.php?title=Q5149038&oldid=1745131299} In later revisions, information about the term's offensiveness was added to the description.}
or the entity Q235155 with the Dutch preferred label \textit{``blanken''} (\textit{``whites''}) and the definition ``term voor mensen met zichtbare Europese oorsprong'' (``term for people with visible European origin'').\footnote{\url{https://www.wikidata.org/w/index.php?title=Q235155&oldid=2005925948}}
Even in these dictionary-like resources, the terms are not always defined as contentious.

As alternative labels, contentious terms are used frequently in Wikidata and AAT.
First, they are used as synonyms of preferred labels, as in the Wikidata entity ``Mumbai'' (Q1156) with the alternative label \textit{``Bombay''}, which is the city's outdated name with colonial connotations.
In this case, contentious alternative labels might be used for discoverability purposes, so that users can find a concept even if they query outdated terms.  
Not all resources, however, contain information about the contentiousness of alternative labels and usually present them as interchangeable synonyms.
Second, an alternative label can be a spelling variation or a word form of a preferred label (which can be both contentious), for example, the AAT concept 300386060 with the Dutch preferred label \textit{``hermafrodieten''} (\textit{``hermaphrodites''}) and the alternative label \textit{``hermafrodiet''} (\textit{``hermaphrodite''}).
In Dutch literals of AAT, there are more preferred labels with contentious terms than alternative.

We checked the pairs of preferred and alternative labels in Wikidata and AAT: whether there were suggestions from the Words Matter KG in alternative labels when contentious terms were used as preferred and vice versa.
In the first case, there is only 1 resource from Wikidata, which is \textit{``Berbers''} (Q45315) with alternative labels \textit{``Berber''}, ``Amazigh'', ``Imazighen''.\footnote{\url{https://www.wikidata.org/w/index.php?title=Q45315&oldid=1822618270}}
The last two terms are suggested to be used instead of \textit{``Berber''} in the Words Matter KG because more people now refer to themselves as Amazigh (while ``Berber'' remains an identity category for some people).
Since the date of retrieval of this entity, its labels were edited several times, and the latest edit placed the term ``Amazigh'' as preferred label.\footnote{\url{https://www.wikidata.org/w/index.php?title=Q45315&oldid=1945487562}}
The second case has 6 resources from Wikidata and 2 from AAT.
For example, the Wikidata entity ``enslaved person'' (Q12773225) had the alternative label \textit{``slave''},\footnote{\url{https://www.wikidata.org/w/index.php?title=Q12773225&oldid=1802607889}} which was also edited since the retrieval, but the latest edit placed the contentious term \textit{``slave''} as preferred label\footnote{\url{https://www.wikidata.org/w/index.php?title=Q12773225&oldid=1950372555}} (in contrast to the example with ``Amazigh'').
These edits illustrate the ongoing discussions about the terms' usage among the Wikidata contributors.

PWN and ODWN, besides lemmas, also frequently mention contentious terms in definitions and examples of synsets.
This is because in both datasets definitions often repeat the term they define and examples illustrate how a term can be used in speech.
In these cases, contentious terms are also often left unexplained in definitions and examples, which reinforces stereotypes.
For example, PWN defines the synset \textit{``mentally\_retarded.n.01''} as \textit{``people collectively who are mentally retarded''}.
And ODWN provides an example for the term \textit{``zigeuner''} (\textit{``gypsy''}) in the synset ``eng-30-10154186-n'': \textit{``er uitzien als een zigeuner''} (\textit{``looking like a gypsy''}).

\subsection{Set 2: All Retrieved Resources}\label{set2}

Querying terms in four datasets resulted in more than 203,000 English and 47,000 Dutch hits.
In Wikidata, we found hits for each query term.
In other datasets, some terms were absent.
Appendix \ref{terms_no_hits} lists terms without hits.

Because of term ambiguity, Set 2 contains literals with terms in contentious meanings, but also other meanings irrelevant to our analysis.
Among the top-10 most frequent terms in all datasets (Appendix \ref{appx_set_1_figures}), we found ambiguous as well as non-ambiguous terms.
Such terms as ``black'' and ``white'', ambiguous in both languages, are frequent across all datasets often denoting colour attributes not related to contentious categories of people's skin colour.
An example of a non-ambiguous term is ``ethnic group'', which occurs in more than 4,000 English and 3,000 Dutch\footnote{There are more than 9,000 entities with Dutch descriptions mentioning the term \textit{``etnische groep''}, but 5,600 of them have missing preferred labels, so it is not possible to determine exactly what these entities denote.} literals of Wikidata entities describing peoples.
There is a separate entity \textit{``ethnic group''} (Q41710) in Wikidata used for categorisation of groups of people.
In AAT, \textit{``ethnic group''} appears in English and Dutch scope notes of more than 400 concepts.
The Words Matter KG explains that the term \textit{``ethnic group''} should be used with caution, because it is usually associated with minority groups and often racialized.

Hits in Set 2 appear in all datasets' property values, except \textit{rdfs:comment} used for Dutch labels in AAT (``Set 2'' in Table \ref{tab:hits_all_levels}).
The majority of the hits is contained in descriptive properties: English and Dutch \textit{skos:scopeNote} of AAT, \textit{schema:description} of Dutch Wikidata, definitions and examples of PWN and ODWN.
There are fewer hits in Dutch literals of Wikidata and AAT.
This can be due to fewer Dutch literals overall, as Table \ref{tab:lod-resources} shows.

Wikidata stands out with a large number of hits in English preferred labels (84,261), where in Dutch, there are 11 times fewer hits.
We found two large categories (values of the ``instance of'' property) of entities only with English preferred labels  without Dutch translation.
These are ``Wikimedia category'' (Q4167836) with 18,100 hits and ``collection'' (Q2668072) with 10,400 hits.
The first is used to categorise articles in Wikipedia and the second represents archival records.
Entities of these two categories frequently mention the terms ``colored'' and ``descent'' in their preferred labels (Figure \ref{fig:wd_en_top10} in Appendix \ref{appx_set_1_figures}).
The term \textit{``colored''} is used in preferred labels of the ``collection'' entities (8,552 hits out of 9,665) representing records about soldiers from the American Civil War period, for example, soldiers from the \textit{``1st US Colored Infantry''}.
These entities were contributed to Wikidata by the U.S. National Archives.
In the ``Wikimedia category'' entities, the term \textit{``descent''} in preferred labels (7,618 hits out of 9,555) denotes people's background, such as \textit{``Category:Canadian people of Chinese descent''} (Q7031448).
Other frequent categories of entities without Dutch preferred labels include titles of literary work, art, and digital media products.

\subsection{Set 3: Resources with Disambiguated Literals}\label{set3}

Disambiguation of all search results produced a subset with 2,307 unique resources.
Almost half of them (1,109) are from Wikidata.
As our evaluation confirmed, 72\% of 240 annotated resources mention terms in a contentious sense.
If we extrapolate this proportion to the whole Set 3, it would result in more than 1,600 unique resources with contentious terms in their literals.

Similar to Set 2, disambiguated literals with contentious terms belong to \textit{skos:scopeNote} of AAT, definitions and examples of PWN and ODWN.
In Wikidata, contentious terms primarily occur in preferred labels: 364 English and 321 Dutch hits (``Set 3'' in Table \ref{tab:hits_all_levels}).

As we did in Set 1, we searched for synonyms of contentious terms in the pairs of preferred and alternative labels of Wikidata and AAT.
In both datasets, there are no other resources (except already included in Set 1) when a contentious term in a preferred label has a suggested synonym from the Words Matter KG in alternative labels.
Only two more resources in Set 3 use suggested terms as preferred when a contentious term is an alternative.
We can conclude that these patterns are rare.

Among the disambiguated literals, we found more resources with contentious preferred labels that are potentially problematic (according to the Words Matter KG): for example, \textit{``Eskimo''} (Q131242), \textit{``Hottentot''} (Q1631241), \textit{``Half-breed''} (Q17144151), \textit{``Mongoloid''} (Q207912), \textit{``gekleurde''} (\textit{``colored''}, about a person) (Q2072081), \textit{``zwarten''} (\textit{``blacks''}, about people) (Q817393) from Wikidata; \textit{``Negro spirituals''} (300393224), \textit{``male homosexuals''} (300435114),  \textit{``Pygmee''} (\textit{``Pygmy''}) (300016430), \textit{``dwergen''} (\textit{``dwarfs''}) (300236748) from AAT.

In AAT, there are significantly more hits in English alternative labels (155) than preferred (58).
This is due to variations of the same label given as alternatives, for example, \textit{``Canadian Eskimo''} and \textit{``Eskimo, Canadian''} (300017455).

PWN and ODWN primarily use contentious terms in definitions and examples of synsets, even when the synsets are not associated with contentious terms.
For example, PWN provides an example sentence \textit{``The gypsies roamed the woods''} for the verb ``roam'' (synset ``roll.v.12'') and the Dutch lemma ``soulmuziek'' (``soul music'') in ODWN is defined as \textit{``muziek onstaan bij negers''} (\textit{``music originating with negroes''}).
In total, there are 370 and 233 potentially problematic definitions and examples in PWN and ODWN, respectively.

\subsection{Markers of Contentiousness}\label{markers}

\begin{table*}
\caption{Types of implicit and explicit markers of contentiousness collected from LOD datasets}
\label{tab:markers_types}
\centering
\resizebox{\textwidth}{!}{%
\begin{tabular}{|l|l|l|}
\hline
\textbf{Type (a marker indicates:)} &
  \textbf{Implicit} &
  \textbf{Explicit} \\ \hline
\#1: offensiveness &
  \begin{tabular}[c]{@{}l@{}}EN: ``offensive'', ``pejorative'', ``derogatory'', ``slur'',\\ ``disparaging term'', ``denigrating'';\\
  NL: ``denigrerend'' (``derogatory''), ``scheldwoord'' (``swear word''),\\ ``negatieve bijklank'' (``negative connotation'')\end{tabular} &
  \begin{tabular}[c]{@{}l@{}}``pejorative'' (Q545779), ``slur'' (Q22116852) in Wikidata;\\ ``Pejorative'' in AAT (\textit{gvp:termKind});\\ ``disparagement.n.01'' in PWN;\\ ``pejorative'' and ``offensive' in ODWN (\textit{Pragmatics connotation})\end{tabular} \\ \hline
\#2: historical usage &
  \begin{tabular}[c]{@{}l@{}}EN: ``obsolete'', ``historical usage'', ``archaic'', ``antiquated'',\\ ``used formerly'', ``older term'';\\
  NL: ``historische'' (``historic''),\\ ``verouderde benaming'' (``obsolete designation'')\end{tabular} &
  \begin{tabular}[c]{@{}l@{}}``historical race concept'' (Q2042898),\\ ``historical profession'' (Q16335296) in Wikidata;\\ the property \textit{gvp:historicFlag} and ``Deprecated'' (\textit{gvp:termKind}) in AAT;\\ ``oldfashioned'' (\textit{Pragmatics chronology}) in ODWN;\end{tabular} \\ \hline
\#3: informal speech &
  \begin{tabular}[c]{@{}l@{}}EN: ``informal'', ``slang'', ``colloquial'';\\ NL: ``informele term'' (``informal term'')\end{tabular} &
  \begin{tabular}[c]{@{}l@{}}``slang.n.02'', ``colloquialism.n.01'' in PWN;\\ ``JargonOrSlang'' (\textit{gvp:termKind}) in AAT\end{tabular} \\ \hline
\#4: (self-)identity categories &
  \begin{tabular}[c]{@{}l@{}}EN: ``term of self-reference'', ``identity term'',\\``self-identifying term'';\\ 
  NL: ``zichzelf aanduiden'' and ``noemen zichzelf'' (``call themselves'')\end{tabular} &
  ``reappropriation'' (Q1520214) in Wikidata \\ \hline
\#5: stereotypes incl. racism &
  \begin{tabular}[c]{@{}l@{}}EN: ``stereotypical'', ``based on stereotypes'', ``ethnic slur'',\\ ``racialized classification'';\\ NL:  ``racistische'' (``racist'')\end{tabular} &
  \begin{tabular}[c]{@{}l@{}}``ethnic slur'' (Q1371427) in Wikidata;\\  ``ethnic\_slur.n.0'' in PWN\end{tabular} \\ \hline
\#6: usage suggestions  &
  \begin{tabular}[c]{@{}l@{}}EN: ``use instead {[}alternative{]}'', ``now prefer {[}alternative{]}'',\\ ``consider'', ``use specific terms'';\\ NL: ``gebruik een specifiekere term'' (``use a more specific term'')\end{tabular} &
  \begin{tabular}[c]{@{}l@{}}``AvoidUse'' (\textit{gvp:termKind}) in AAT;\\ literals of ``Wikidata usage instructions'' (P2559) in Wikidata;\end{tabular} \\ \hline
\end{tabular}%
}
\end{table*}

\paragraph{Implicit Markers}

We identified 31 resources with implicit markers in Set 1 out of 261 unique resources (12\%).
Most of these resources come from Wikidata (12); no implicit markers were found in ODWN.
Implicit markers were given via descriptive properties: \textit{schema:description} in Wikidata, synset definitions in PWN, \textit{skos:scopeNote} and \textit{rdfs:comments} 
in AAT.
Querying literals of these properties, we retrieved 74 and 20 resources more with implicit markers from Sets 3 and 2, respectively.
Most implicit markers are only in English.
In Dutch, they occurred in 19 resources of all sets.

Depending on what implicit markers indicate, we categorised them by 6 types (Table \ref{tab:markers_types}) .
In almost a third of the cases, implicit markers indicate that a term carries offensive meaning.
For example, PWN defines the term \textit{``coolie''} as ``(ethnic slur) an offensive name for an unskilled Asian laborer'' in the synset \textit{coolie.n.01}.
In some resources, AAT contained implicit markers in long texts of scope notes with explanations and suggestions on terms' usage (Type \#6).
For example, the concept \textit{``Eskimo (culture or style)''} (300017447) has a scope note ``For names of specific native peoples of the present, use descriptors such as "Chugach," "Inuit," or "Katladlit."''.
These suggestions are similar to those given in the Words Matter KG.

\paragraph{Explicit Markers}

36 resources in Set 1 (14\%) contained properties, relations to other resources, and tags explicitly indicating contentiousness.
Furthermore, we retrieved 121 resources in Set 2 and 28 resources in Set 3 by querying the identified properties and tags used for explicit marking.

Comparable to implicit markers, explicit markers also come in different types (Table \ref{tab:markers_types}).
In Wikidata, several entities are used to mark contentiousness.
For example, the entity \textit{``mulatto''} (Q191923) is an ``instance of'' (P31) ``historical race concept'' (Q2042898).
The Wikidata property ``Wikidata usage instructions'' (P2559) connects entities to information on how contentious terms should be used.
AAT flags some preferred and alternative labels with the properties \textit{gvp:termKind} and \textit{gvp:historicFlag}.
The values of \textit{gvp:termKind} are, for example, ``AvoidUse'', ``JargonOrSlang'', or ``Pejorative''.
And \textit{gvp:historicFlag} has only two values ``historic'' and ``currentAndHistoric''.
The synsets of PWN are connected via \textit{usage\_domain} to other synsets, for example, \textit{disparagement.n.01} or \textit{slang.n.02}.
The ODWN tag \textit{Pragmatics} has several attributes, including ``connotation'' and ``chronology'' with values indicating offensiveness (``pejorative'', ``offensive'') or that a term is ``oldfashioned''.

By querying explicit markers outside of the retrieved resources, we found that the same markers were used for purposes other than signalling contentiousness.
For example, ``Wikidata usage instructions'' (P2559) explains technical details on how qualifiers should be used.
In AAT, the property \textit{gvp:termKind} also has values not related to contentiousness, such as ``Misspelling'' or ``Abbreviation''.
And the value ``historic'' of \textit{gvp:historicFlag} also marks historical spelling of terms.
The \textit{usage\_domain} in PWN indicates, besides offensive lemmas, comparative and plural forms.
Such weak semantics of these properties complicate the identification of resources with contentious terms in their literals.

\paragraph{Resources Without Markers}

In Set 3, we checked whether contentious terms in preferred labels of Wikidata and AAT resources and in lemmas of PWN and ODWN synsets had any markers.
Resources are rarely marked either implicitly or explicitly in all datasets.
Out of 657 Wikidata resources with disambiguated preferred labels, only 21 were marked.
In AAT, this proportion is 91/14.
For example, the Wikidata entity \textit{``Pygmy people''} (Q171927) does not have any markers, while in AAT, the concept \textit{``Pygmy (African culture or style)''} (300016430) has a scope note with an implicit marker stating that ``Use of \textit{"Pygmy"} for a culture is considered pejorative''.
At the same time, this AAT concept has the Dutch preferred label \textit{``Pygmee''} (\textit{``Pygmy''}), which is not marked.
This illustrates such cases where even if a contentious term marked as ``pejorative'', it still can be used as a preferred label and shown to users in interfaces that, for example, ignore scope notes.

In PWN, there are 117 disambiguated lemmas with contentious terms, and 11 of them were marked.
ODWN marked 10 lexical entries out of 79.
The definition of the PWN synset ``black.n.05'' marks the terms \textit{``Negro''} and \textit{``Negroid''} as ``archaic and pejorative today'', however the synset ``colored.s.02'' lists the term \textit{``negro''} as a synonym of \textit{``colored''} and \textit{``coloured''} without markers.
The tag ``Pragmatics connotation'' in ODWN marks the lemma``flikker''(\textit{``faggot''}) and its synonyms as ``pejorative'' and ``offensive'' (synset ``eng-30-10076033-n''), but the synset ``eng-30-10182913-n'' with the same lemma (with a different id) does not contain any markers.
It results in the situation when the words from this synset are being used in interfaces of online dictionaries,\footnote{\url{https://babelnet.org/synset?id=bn\%3A00037547n\&orig=homoseksuele\&lang=NL}. Last accessed on 13.11.2023} which display offensive terms \textit{``bruinwerker''} (literally \textit{``brown worker''}), \textit{``flikker''} (\textit{``faggot''}), and \textit{``geïnverteerde''} (\textit{``inverted''}) as regular synonyms of \textit{``homosexuality''}.

%% file: src/conclusion.tex
Contentious terms that express stereotypes about people and cultures and which are potentially offensive to users appear on a large scale in four widely used linked open datasets, our study showed.
We extracted and analysed the literals mentioning contentious terms from Wikidata, The Getty Art \& Architecture Thesaurus (AAT) common in the cultural sector, and two lexical databases Princeton WordNet (PWN) and Open Dutch WordNet (ODWN).
The terms we studied originate from a knowledge graph, which models judgements of experts from the cultural heritage domain about terms' cultural sensitivities.

From the four datasets, we retrieved more than 203,000 English and 47,000 Dutch literals.
Because of term ambiguity, there were literals with terms in non-contentious senses.
Applying word sense disambiguation, we constructed a subset with a large proportion of relevant literals, which still contains more than {2,000} resources.

In the set with disambiguated literals, Wikidata has the largest number of resources using both English and Dutch contentious terms as preferred labels.
We found preferred labels of Wikidata resources, which refer to people in an outdated and derogatory fashion without (more appropriate) synonyms in alternative labels or explanations about the terms' usage in descriptions.
In other datasets, contentious terms are primarily used in descriptive literals: scope notes of AAT, definitions and examples of PWN and ODWN.

Our analysis revealed various attempts of the LOD contributors marking resources with contentious terminology: using special words and phrases in literals (implicit markers) or properties (explicit markers).
Implicit markers are not frequent and appear mostly in English literals of descriptive properties.
There are language independent properties used for explicit marking in each dataset, however, some of them have weak semantics: they are used for multiple purposes other than signalling contentiousness.
In all datasets, we found marking rare and not systematic.

The insights we gathered can serve as a starting point towards compiling more informative guidelines for those who describe data and designing approaches to detect and prevent the propagation of stereotypes on the Web more systematically.
\balance

%% file: src/appendix.tex
\newpage
\section{Terms Without Related Resources}\label{terms_no_related_resources}

{Wikidata EN (15): \textit{``baboo''}, \textit {``bush negro''}, \textit {``developing nations''}, \textit {``discover''}, \textit {``dwarf''}, \textit {``exotic''}, \textit {``full blood''}, \textit {``half-blood''}, \textit {``headhunter''}, \textit {``homo''}, \textit {``low-income countries''}, \textit {``native''}, \textit {``oriental''}, \textit {``roots''}, \textit {``traditional''}.

Wikidata NL (13): \textit{``achterlijk''}, \textit{``exotisch''}, \textit{``halfbloed''}, \textit{``inboorling''}, \textit{``knecht''}, \textit{``koppensneller''}, \textit{``ontdekken''}, \textit{``oriëntaals''}, \textit{``primitief''}, \textit{``roots''}, \textit{``traditioneel''}, \textit{``trans''}, \textit{``volbloed''}.

AAT EN (34): \textit{``allochtoon''}, \textit{``baboo''}, \textit{``barbarian''}, \textit{``batavia''}, \textit{``bombay''}, \textit{``burma''}, \textit{``bush negro''}, \textit{``calcutta''}, \textit{``colored''}, \textit{``coolie''}, \textit{``descent''}, \textit{``discover''}, \textit{``exotic''}, \textit{``footmen''}, \textit{``full blood''}, \textit{``half-blood''}, \textit{``half-breed''}, \textit{``handicap''}, \textit{``headhunter''}, \textit{``homo''}, \textit{``kaffir''}, \textit{``lilliputian''}, \textit{``low-income countries''}, \textit{``madras''}, \textit{``mestizo''}, \textit{``mongoloid''}, \textit{``mulatto''}, \textit{``page''}, \textit{``roots''}, \textit{``second world''}, \textit{``southern rhodesia''}, \textit{``trans''}, \textit{``western''}, \textit{``white''}.

AAT NL (55): \textit{``afkomst''}, \textit{``allochtoon''}, \textit{``baboe''}, \textit{``barbaar''}, \textit{``batavia''}, \textit{``birma''}, \textit{``blank''}, \textit{``bombay''}, \textit{``boslandcreool''}, \textit{``bosneger''}, \textit{``calcutta''}, \textit{``derde wereld''}, \textit{``eerste wereld''}, \textit{``exotisch''}, \textit{``gay''}, \textit{``gekleurd''}, \textit{``halfbloed''}, \textit{``handicap''}, \textit{``homo''}, \textit{``homoseksueel''}, \textit{``hottentot''}, \textit{``inboorling''}, \textit{``inlander''}, \textit{``islamiet''}, \textit{``jap''}, \textit{``jappenkamp''}, \textit{``kaffer''}, \textit{``knecht''}, \textit{``koelie''}, \textit{``koppensneller''}, \textit{``lagelonenland''}, \textit{``lilliputter''}, \textit{``madras''}, \textit{``marron''}, \textit{``medicijnman''}, \textit{``mesties''}, \textit{``mohammedaan''}, \textit{``mulat''}, \textit{``neger''}, \textit{``ontdekken''}, \textit{``oriëntaals''}, \textit{``page''}, \textit{``politionele actie''}, \textit{``primitief''}, \textit{``queer''}, \textit{``roots''}, \textit{``traditioneel''}, \textit{``trans''}, \textit{``travestiet''}, \textit{``tweede wereld''}, \textit{``volbloed''}, \textit{``westers''}, \textit{``wit''}, \textit{``zuid-rhodesië''}, \textit{``zwart''}.

PWN (20): \textit{``allochtoon''}, \textit{``baboo''}, \textit{``batavia''}, \textit{``bush negro''}, \textit{``developing nations''}, \textit{``ethnic groups''}, \textit{``first world''}, \textit{``full blood''}, \textit{``half-blood''}, \textit{``hottentot''}, \textit{``indo''}, \textit{``low-income countries''}, \textit{``maroon''}, \textit{``medicine man''}, \textit{``métis''}, \textit{``roots''}, \textit{``second world''}, \textit{``southern rhodesia''}, \textit{``third world''}, \textit{``trans''}.

ODWN (25): \textit{``batavia''}, \textit{``boslandcreool''}, \textit{``derde wereld''}, \textit{``eerste wereld''}, \textit{``etnische groep''}, \textit{``exotisch''}, \textit{``gekleurd''}, \textit{``indisch''}, \textit{``indo''}, \textit{``inheems''}, \textit{``lagelonenland''}, \textit{``marron''}, \textit{``métis''}, \textit{``oriëntaals''}, \textit{``politionele actie''}, \textit{``primitivisme''}, \textit{``queer''}, \textit{``ras''}, \textit{``roots''}, \textit{``traditioneel''}, \textit{``tweede wereld''}, \textit{``volbloed''}, \textit{``westers''}, \textit{``wit''}, \textit{``zuid-rhodesië''}.

All datasets EN (6): \textit{``baboo''}, \textit{``bush negro''}, \textit{``full blood''}, \textit{``half-blood''}, \textit{``low-income countries''}, \textit{``roots''}.

All datasets NL (5): \textit{``exotisch''}, \textit{``oriëntaals''}, \textit{``roots''}, \textit{``traditioneel''}, \textit{``volbloed''}.

\section{Terms Without Search Hits}\label{terms_no_hits}

AAT EN (11): \textit{``allochtoon''}, \textit{``baboo''}, \textit{``bush negro''}, \textit{``coolie''}, \textit{``full blood''}, \textit{``half-blood''}, \textit{``half-breed''}, \textit{``headhunter''}, \textit{``lilliputian''}, \textit{``low-income countries''}, \textit{``southern rhodesia''}.

AAT NL (26): \textit{``allochtoon''}, \textit{``baboe''}, \textit{``barbaar''}, \textit{``boslandcreool''}, \textit{``bosneger''}, \textit{``eerste wereld''}, \textit{``halfbloed''}, \textit{``inboorling''}, \textit{``inlander''}, \textit{``jap''}, \textit{``jappenkamp''}, \textit{``kaffer''}, \textit{``koelie''}, \textit{``koppensneller''}, \textit{``lagelonenland''}, \textit{``lilliputter''}, \textit{``marron''}, \textit{``medicijnman''}, \textit{``mesties''}, \textit{``neger''}, \textit{``politionele actie''}, \textit{``queer''}, \textit{``roots''}, \textit{``tweede wereld''}, \textit{``volbloed''}, \textit{``zuid-rhodesië''}.

PWN (7): \textit{``allochtoon''}, \textit{``batavia''}, \textit{``bush negro''}, \textit{``full blood''}, \textit{``half-blood''}, \textit{``low-income countries''}, \textit{``southern rhodesia''}.

ODWN (11): \textit{``boslandcreool''}, \textit{``eerste wereld''}, \textit{``lagelonenland''}, \textit{``marron''}, \textit{``métis''}, \textit{``oriëntaals''}, \textit{``primitivisme''}, \textit{``queer''}, \textit{``roots''}, \textit{``tweede wereld''}, \textit{``zuid-rhodesië''}.

\section{The Most Frequent Terms in Set 1}\label{appx_set_1_figures}

The stacked bar charts illustrate the proportion of the property values, in which contentious terms occur.
The total number of hits per term is shown on the right.
\balance
\newpage
    
\begin{figure}[H]
    \centering
    \includegraphics[width=\linewidth]{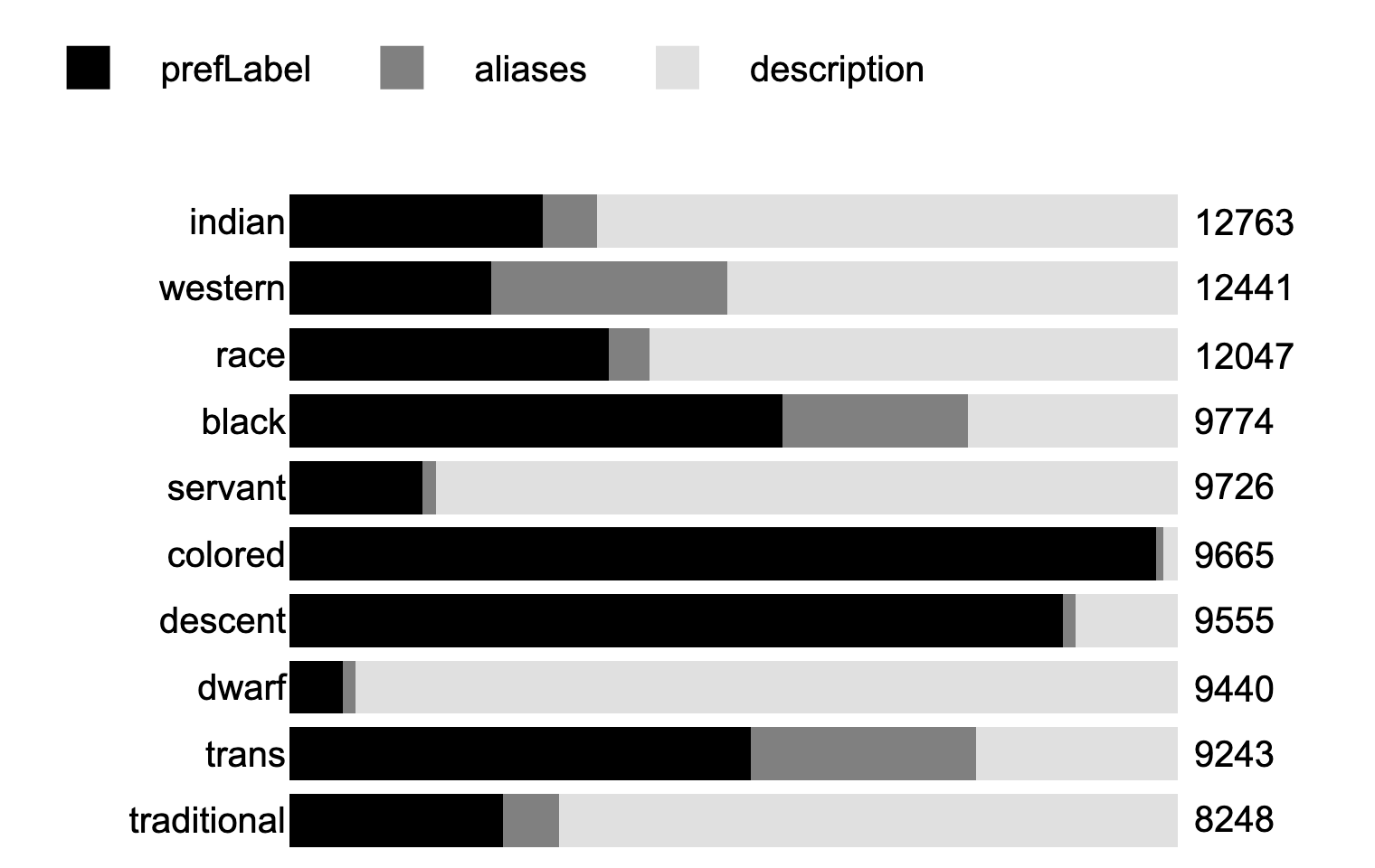}
    \caption{Top-10 EN contentious terms by N hits in Wikidata}
    \label{fig:wd_en_top10}
\end{figure}

\begin{figure}[H]
\centering
    \includegraphics[width=\linewidth]{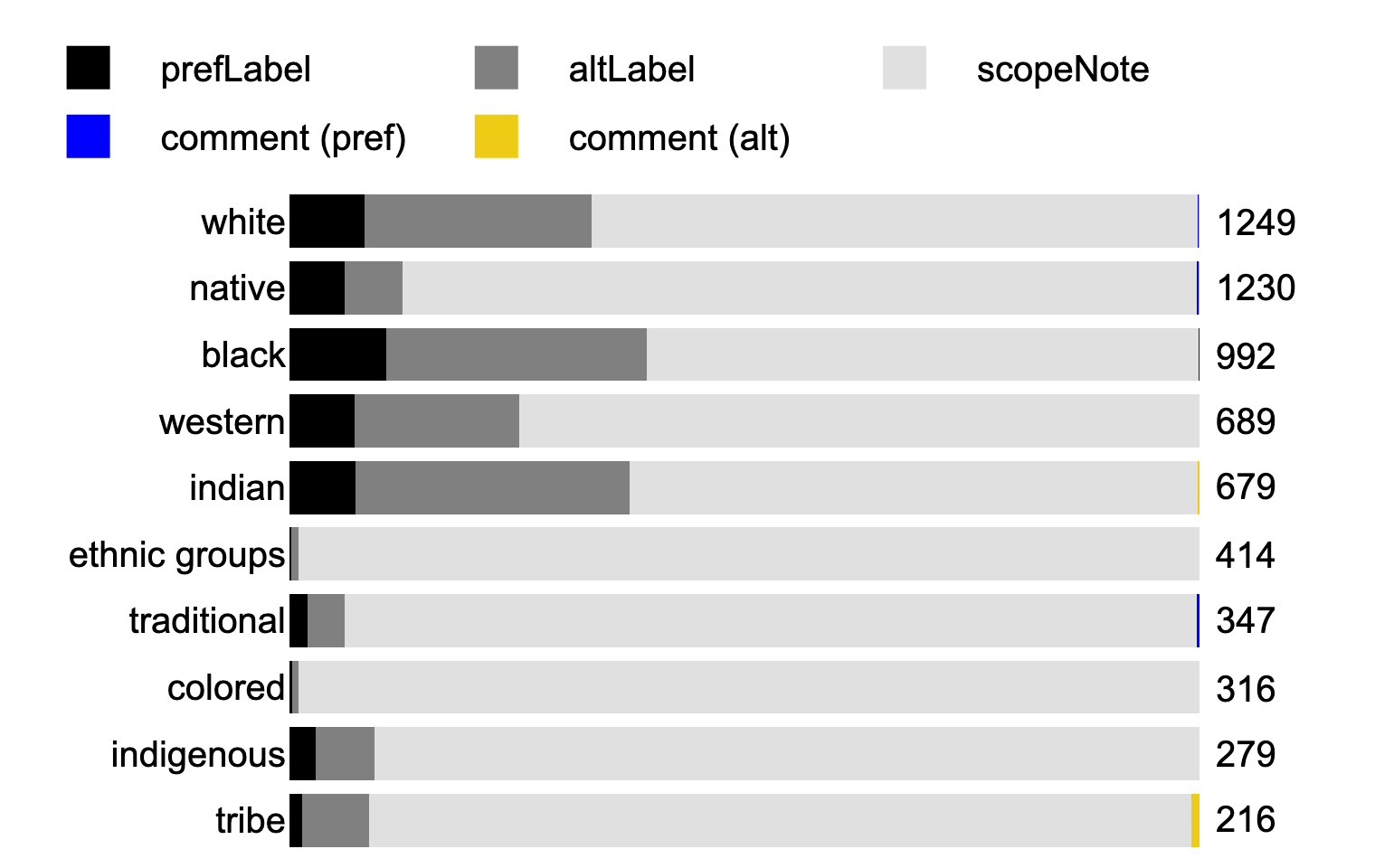}
    \caption{Top-10 EN contentious terms by N hits in AAT}
    \label{fig:aat_en_top10}
\end{figure}

\begin{figure}[H]
    \centering
    \includegraphics[width=\linewidth]{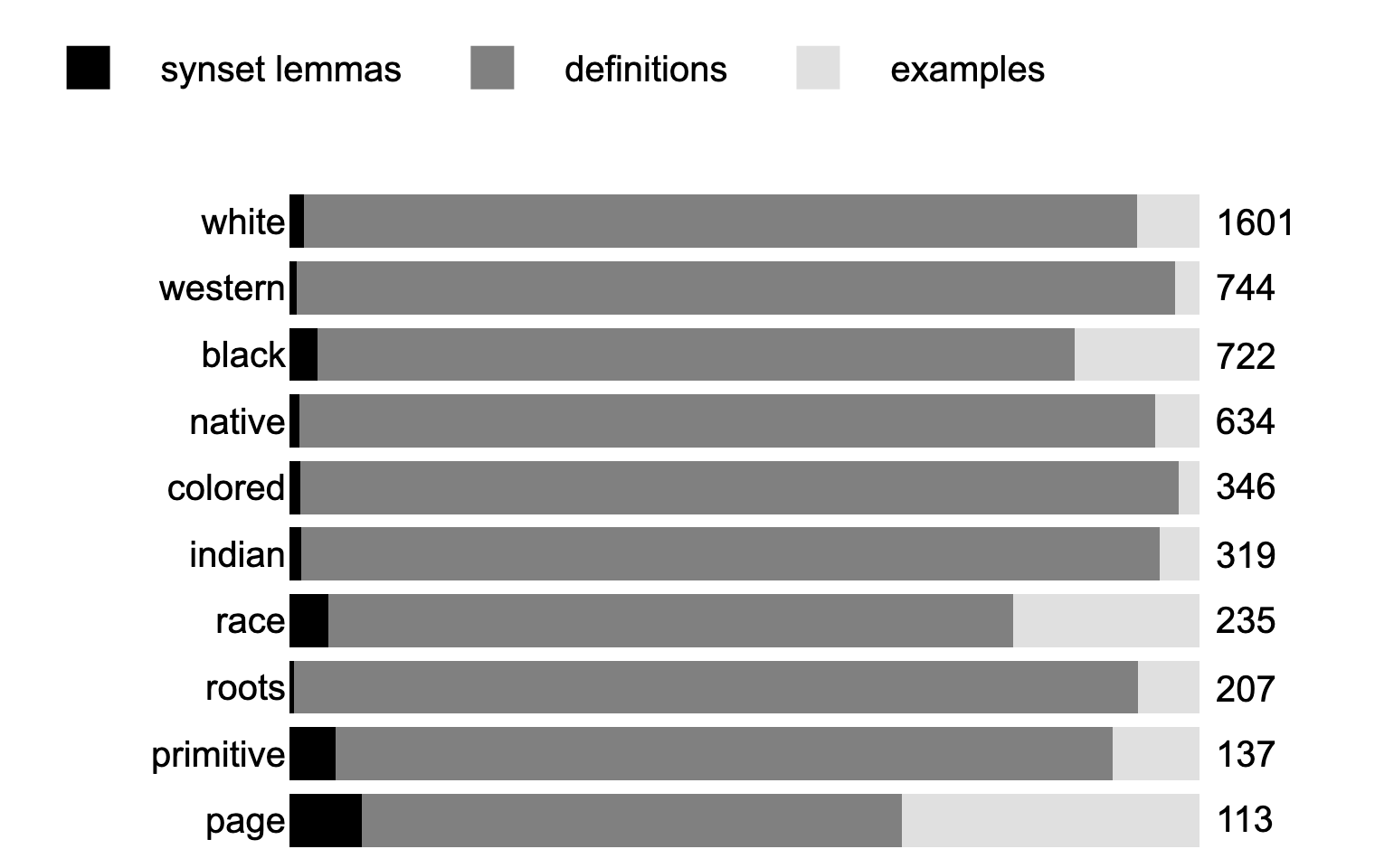}
    \caption{Top-10 EN contentious terms by N hits in PWN}
    \label{fig:pwn_top10}
\end{figure}


\begin{figure}[H]
    \centering
    \includegraphics[width=\linewidth]{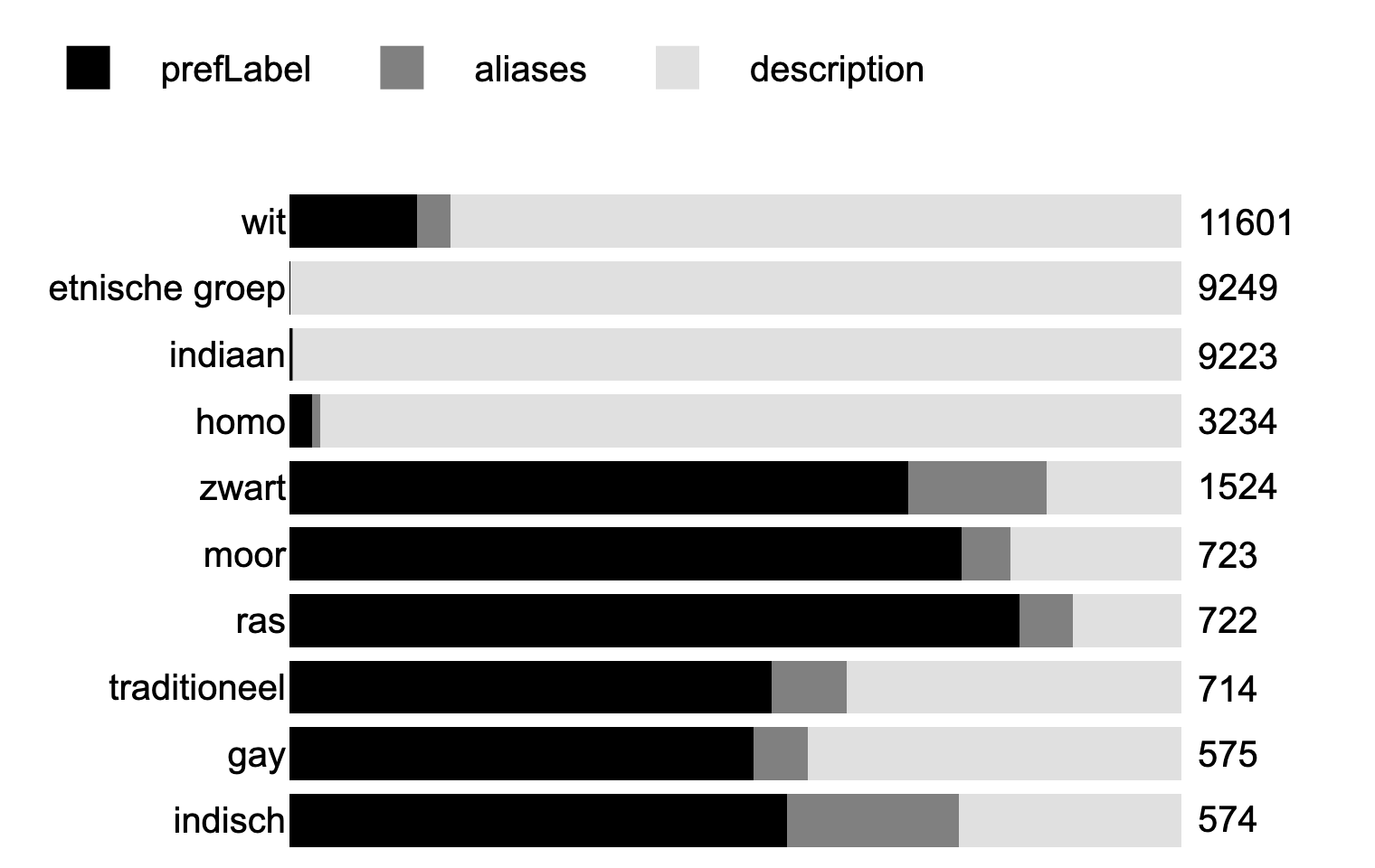}
    \caption{Top-10 NL contentious terms by N hits in Wikidata}
    \label{fig:wd_nl_top10}
\end{figure}

\begin{figure}[h]
    \centering
    \includegraphics[width=\linewidth]{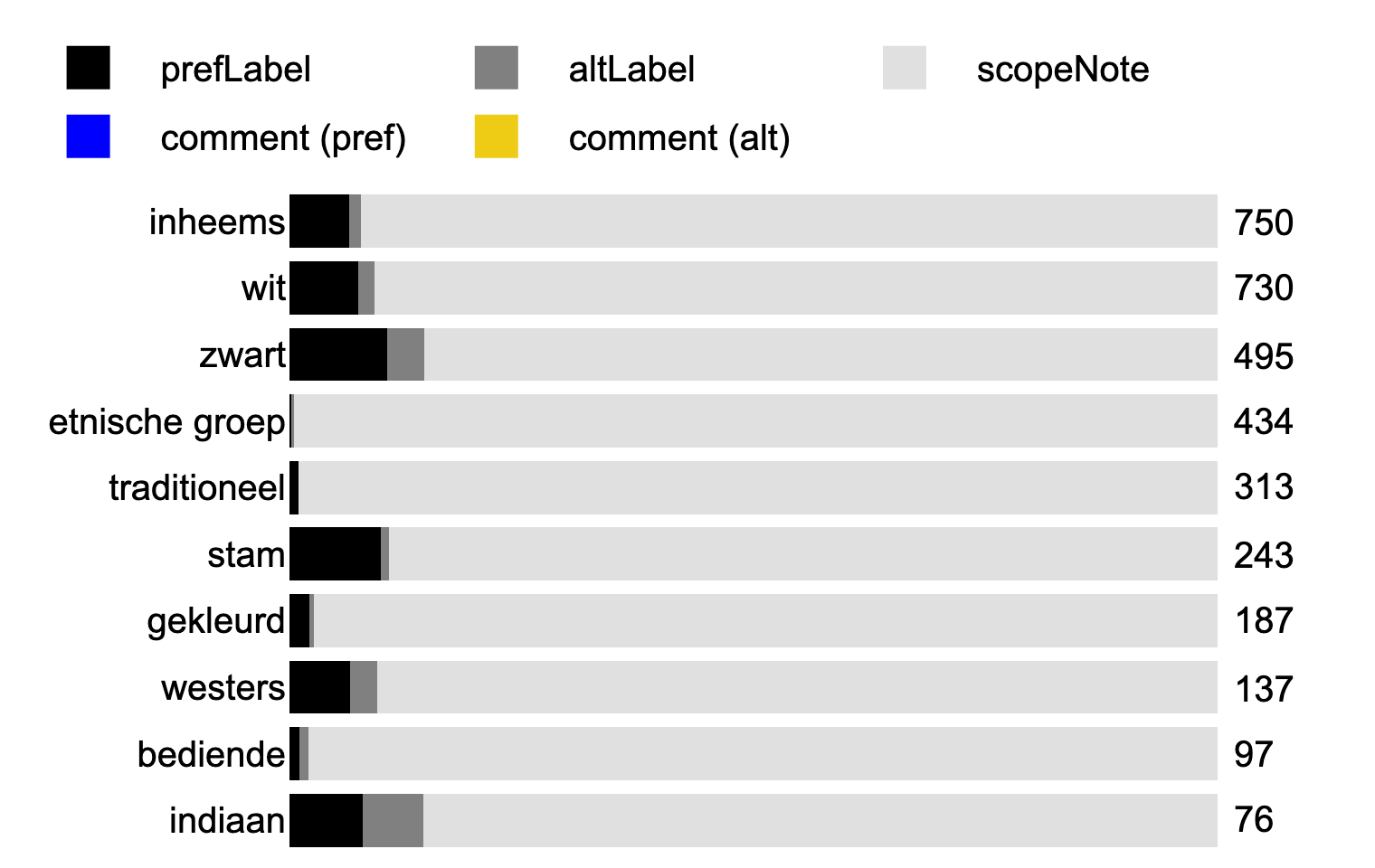}
    \caption{Top-10 NL contentious terms by N hits in AAT}
    \label{fig:aat_nl_top10}
\end{figure}

\begin{figure}[h]
    \centering
    \includegraphics[width=\linewidth]{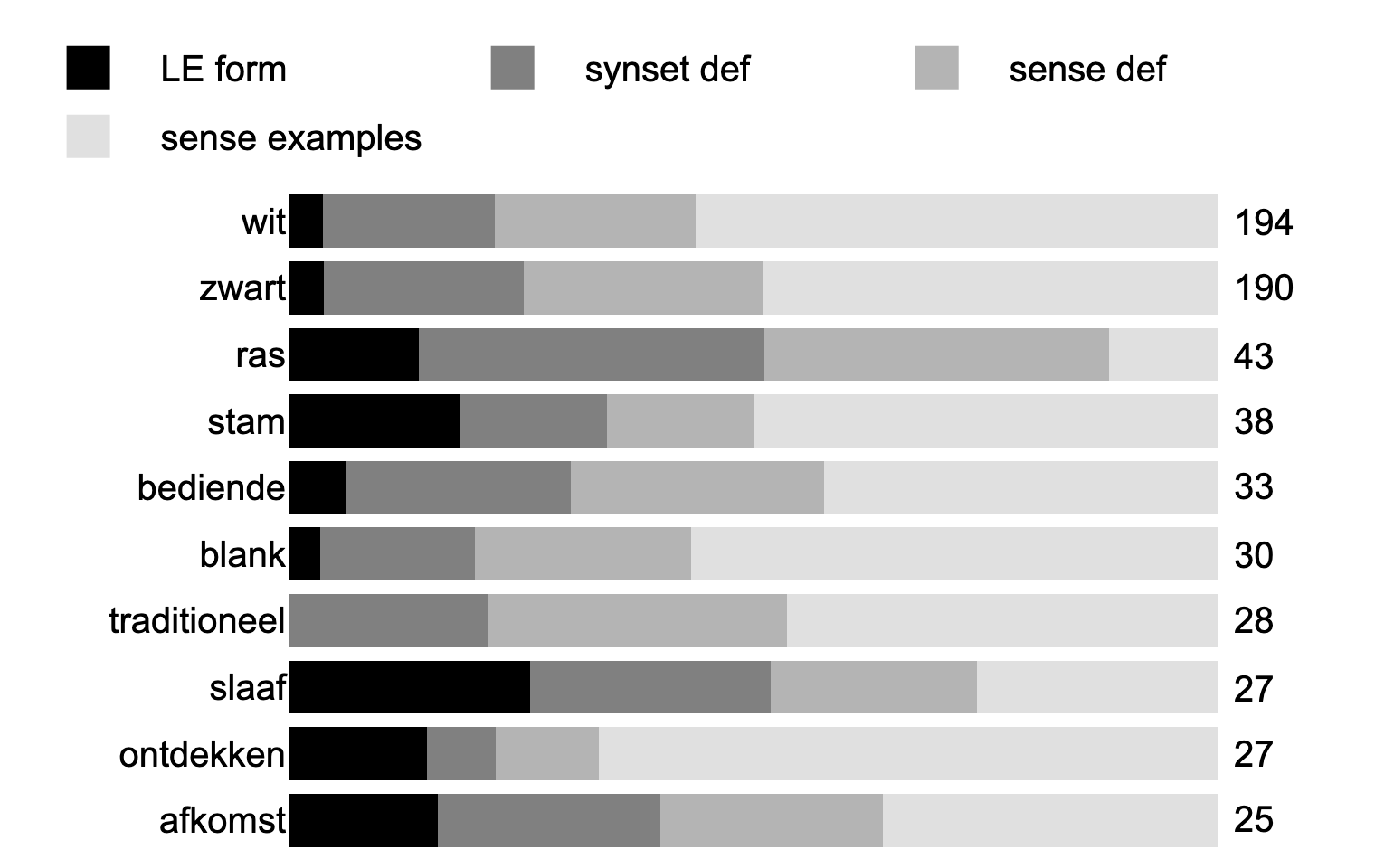}
    \caption{Top-10 NL contentious terms by N hits in ODWN}
    \label{fig:odwn_top10}
\end{figure}